\documentclass[11pt]{article}
\usepackage{mathpazo}
\usepackage{fullpage}

\usepackage{hyperref}       
\PassOptionsToPackage{hyphens}{url}
\usepackage{url}            
\usepackage{booktabs}       
\usepackage{amsfonts}       
\usepackage{amsmath}
\usepackage{cleveref}
\usepackage{nicefrac}       
\usepackage{microtype}      
\usepackage{graphicx} 
\usepackage{wrapfig}
\usepackage{enumerate}
\usepackage{tikz,tikz-qtree}
\usepackage{import}
\usepackage{enumitem}
\usepackage{subcaption}
\usepackage[round, numbers]{natbib}
\usepackage{comment}

\usepackage{xcolor}
\newcommand{\Micah}[1]{{\textcolor{blue}{\textbf{Micah:} #1 }}}
\newcommand{\xinyun}[1]{{\textcolor{red}{\textbf{Xinyun:} #1 }}}

\usepackage[textsize=scriptsize, textwidth=1.3cm]{todonotes}

\DeclareMathOperator*{\argmin}{argmin}
\DeclareMathOperator*{\st}{subject\,\, to}

\renewcommand{\paragraph}[1]{\noindent\textbf{#1}}

\usepackage{xspace}

\makeatletter
\newcommand{\citepp}[1]{(\citeauthor{#1} \citeyear{#1}\checknextarg}
\newcommand{\checknextarg}{\@ifnextchar\bgroup{\gobblenextarg}{)\xspace}}
\newcommand{\gobblenextarg}[1]{, \citeauthor{#1} \citeyear{#1}\@ifnextchar\bgroup{\gobblenextarg}{)\xspace}}

\newcommand{\citett}[1]{\citeauthor{#1}~(\citeyear{#1})\xspace}

\usepackage{wrapfig}

\usepackage{authblk}

\usepackage[T1]{fontenc}

\newcommand{\eat}[1]{}

\begin{document}



\title{Dataset Security for Machine Learning: \\Data Poisoning, Backdoor Attacks, and Defenses}
\date{}
\author[1]{Micah Goldblum\footnote{\href{goldblum@umd.edu}{goldblum@umd.edu}}}
\author[2]{Dimitris Tsipras}
\author[3]{Chulin Xie}
\author[4]{Xinyun Chen}
\author[1]{Avi Schwarzschild}
\author[4]{\\Dawn Song}
\author[2]{Aleksander M\k{a}dry}
\author[3]{Bo Li}
\author[1]{Tom Goldstein\footnote{\href{tomg@cs.umd.edu}{tomg@cs.umd.edu}}}
\affil[1]{University of Maryland}
\affil[2]{Massachussetts Institute of Technology}
\affil[3]{University of Illinois Urbana-Champaign}
\affil[4]{University of California, Berkeley}
\setcounter{Maxaffil}{0}
\renewcommand\Affilfont{\itshape\small}


\maketitle

\begin{abstract}
As machine learning systems grow in scale, so do their training data requirements, forcing practitioners to automate and outsource the curation of training data in order to achieve state-of-the-art performance.
 The absence of trustworthy human supervision over the data collection process exposes organizations to security vulnerabilities; training data can be manipulated to control and degrade the downstream behaviors of learned models.  
The goal of this work is to systematically categorize and discuss a wide range of dataset vulnerabilities and exploits, approaches for defending against these threats, and an array of open problems in this space.  In addition to describing various poisoning and backdoor threat models and the relationships among them, we develop their unified taxonomy.
\end{abstract}

\section{Introduction}
\label{sec:intro}


Traditional approaches to computer security isolate systems from the outside world through a combination of firewalls, passwords, data encryption, and other access control measures.  In contrast, dataset creators often invite the outside world in --- data-hungry neural network models are built by harvesting information from anonymous and unverified sources on the web.  
%
%
Such open-world dataset creation methods can be exploited in several ways.   Outsiders can {\em passively} manipulate datasets by placing corrupted data on the web and waiting for data harvesting bots to collect them.    {\em Active} dataset manipulation occurs when outsiders have the privilege of sending corrupted samples directly to a dataset aggregator such as a chatbot, spam filter, or database of user profiles.  Adversaries may also inject data into systems that rely on {\em federated learning}, in which models are trained on a diffuse network of edge devices that communicate periodically with a central server. In this case, users have complete control over the training data and labels seen by their device, in addition to the content of updates sent to the central server. 
 The exploitability of web-based dataset creation is illustrated by the manipulation of the Tay chatbot \citepp{wakefield2016microsoft}, the presence of potential malware embedded in ImageNet files \citepp{reddit2020}, and manipulation of commercial spam filters \citepp{nelson2008exploiting}.  A recent survey of industry practitioners found that organizations are significantly more afraid of data poisoning than other threats of adversarial machine learning \citepp{kumar2020adversarial}.


The goal of this article is to catalog and systematize vulnerabilities in the dataset creation process, and review how these weaknesses lead to exploitation of machine learning systems.  We will address the following dataset security issues: 
\begin{itemize}[leftmargin=5mm]
\item {\em Training-only attacks: } These attacks 
entail manipulating training data and/or labels and require no access to test-time inputs after a system is deployed.  They can be further grouped by the optimization formulation or heuristic used to craft the attack and whether they target a from-scratch training process or a transfer learning process that fine-tunes a pre-trained model.
\item {\em Attacks on both training and testing: } These threats are often referred to as ``backdoor attacks'' or ``trojans.''  They embed an exploit at train time that is subsequently invoked by the presence of a ``trigger'' at test time.  These attacks can be further sub-divided into model-agnostic attacks and model-specific attacks that exploit a particular neural network architecture.  Additional categories of attacks exist that exploit special properties of the transfer learning and federated learning settings.
\item {\em Defenses against dataset tampering: }  Defense methods can either detect when poisoning has taken place or produce an unaffected model using a training process that resists poisons.   Detection methods include those that identify corrupted training instances, in addition to methods for flagging corrupted models after they have been trained.  Training-based defenses may avoid the consequences of poisoning altogether using robust training routines, or else perform post-hoc correction of a corrupted model to remove the effects of poisoning.
\end{itemize}

In our treatment, we also discuss various threat models addressed in the literature and how they differ from each other.  Finally, for each of the three topics above, we discuss open problems that, if solved, would increase our understanding of the severity of a given class of attacks, or enhance our ability to defend against them.

Unlike data poisoning and backdoor attacks, evasion attacks act during inference.  For a survey on evasion attacks, see \citett{yuan2019adversarial}.  A more focused treatment of threats against federated learning is found in \citett{lyu2020threats}.  Similarly, \citett{li2020backdoor} and \citett{gao2020backdoor} focus on backdoor threats.  A broad discussion concerning security for machine learning can be found in \citett{papernot2018sok} and \citett{pitropakis2019taxonomy}.

\section{Training-Only Attacks}
\label{sec:train}
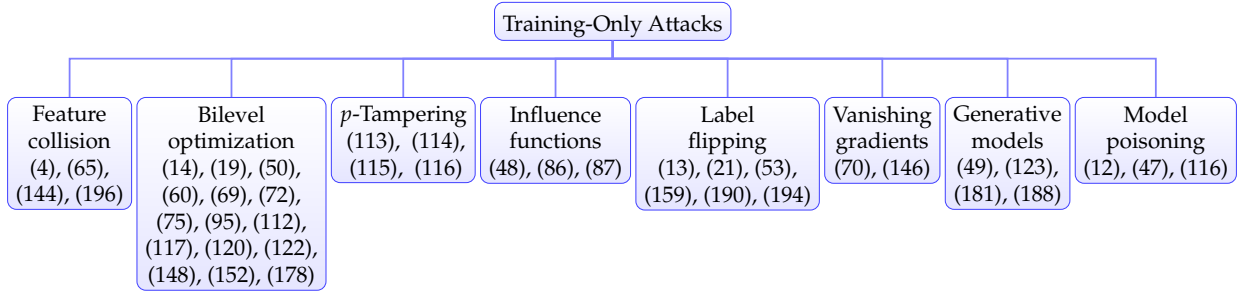
\begin{figure*}[h!]
    \centering
    \resizebox{\textwidth}{!}{
        \begin{tikzpicture}[
            i/.style={
                anchor=north,
                align=center,
                top color=white,
                bottom color=blue!10,
                rectangle,rounded corners,
                minimum height=6mm,
                draw=blue!75,
                align=center,
                color=blue!75,
                text=black,
                text depth = 0pt
            }
        ]
            \tikzset{execute at begin node=\strut}
            \tikzset{font=\small,
                grow=down,
                level distance=1.5cm,
                every tree node/.style={align=center,anchor=north},
                every leaf node/.style=
                    {
                        anchor=north,
                        align=center,
                        fill=blue!5,
                        rectangle,
                        draw=blue!50,
                        align=center,
                        distance=1cm,
                        text depth = -10pt
                    },
                edge from parent/.style=
                    {
                        draw=blue!50,
                        thick,
                        edge from parent path={(\tikzparentnode.south)
                        -- +(0,-8pt)
                        -| (\tikzchildnode)}
                    }
            }

            \Tree [.\node[i] {Training-Only Attacks};
            [
                .\node[i] {Feature \\ collision\\
                \cite{aghakhani2020bullseye},
                   \cite{guopractical},   \\
                   \cite{shafahi2018poison}, 
                   \cite{zhu2019transferable}};
            ]
            [
                .\node[i] {Bilevel \\ optimization\\
                  \cite{biggio2012poisoning}, 
                   \cite{burkard2017analysis},
                   \cite{fowl2021preventing}, \\
                   \cite{geiping2020witches}, 
                   \cite{hu2019targeted}, 
                   \cite{huang2020metapoison}, \\
                   \cite{jagielski2018manipulating},
                   \cite{li2016data},
                   \cite{ma2019dplearners}, \\
                   \cite{mei2015using},                      
                   \cite{miao2018towards}, 
                   \cite{munoz2017towards}, \\
                   \cite{solans2020poisoning},
                   \cite{sun2020data},
                   \cite{xiao2015feature}};
            ]
            [.\node[i]{$p$-Tampering\\
                                ~\cite{mahloujifar2017blockwise},
                    ~\cite{mahloujifar2018learning}, \\
                    ~\cite{mahloujifar2019curse},
                    ~\cite{mahloujifar2019data}};
            ]
            [.\node[i]{Influence \\ functions\\
            \cite{fang2020influence},
                    \cite{koh2017influencefunction},
                    \cite{koh2018stronger}};
            ]
            [.\node[i]{Label \\ flipping\\
                    \cite{biggio2011support},
                    \cite{cao2019understanding},
                    \cite{fung2018mitigating}, \\
                    \cite{tolpegin2020data},
                    \cite{zhang2017game},
                    \cite{zhao2017efficient}};
            ]
            [.\node[i]{Vanishing \\ gradients\\
            \cite{huang2021unlearnable},
            \cite{shen2019tensorclog}};
            ]
            [.\node[i]{Generative \\ models\\
            \cite{feng2019learning},
            \cite{munoz2019poisoning},\\
                    \cite{yang2017generative},
                    \cite{zhang2019poisoning}};
            ]
            [.\node[i]{Model \\ poisoning\\
            \cite{bhagoji2019analyzing},
                    \cite{fang2020local},
                    \cite{mahloujifar2019data}};
            ]
            ]
        \end{tikzpicture}
    }
    \caption{A taxonomy of training-only data poisoning attacks.}
    \label{fig:training-only-tree}
\end{figure*}
A number of data poisoning strategies manipulate training data without the need to modify test instances in the field after the victim model is deployed. Training-only attacks are salient in scenarios where training data is collected from potentially compromised online sources.  Such sources include social media profiles, where users can manipulate text or embed exploits in images scraped for facial recognition.  Figure \ref{fig:training-only-tree} contains a visual depiction of the taxonomy of training-only attacks based on their methodologies.

\subsection{Applications of Data Poisoning}
The broad range of applications of training-only attacks includes both \emph{targeted attacks} in which the attacker seeks to change the behavior of the model on particular inputs or individuals, and \emph{untargeted attacks} where the attacker's impact is meant to indiscriminately affect model behavior on a wide range of inputs. 
An example of a targeted poisoning attack is Venomave \citepp{aghakhani2020venomave}, which attacks automatic speech recognition systems to alter the model's classification of a particular person's utterance of a specific numerical digit.  An example of an untargeted attack is one that reduces algorithmic fairness at the population level~\citepp{solans2020poisoning}.  

Poisoning attacks reveal vulnerabilities not only in neural networks, but in simple classical models as well. For example, the spam filtering algorithm SpamBayes \citepp{meyer2004spambayes}, which uses a naive Bayes classifier to filter email, is susceptible to poisoning attacks \citepp{nelson2008exploiting}. By including many words from legitimate emails in messages labeled as spam in the training set, the attacker can increase the spam score on legitimate emails at test time. When the attacker has access to a sample of the victim's legitimate email, the attacker can use the distribution of the words from that sample to craft undetectable spam. Without such access, the attacker can include samples from a dictionary of words associated with legitimate email or spam \citepp{nelson2008exploiting}.

\textbf{Recommendation systems.} One well-studied application of data poisoning is recommendation systems \citepp{li2016data}{fang2018poisoning}{hu2019targeted}{fang2020influence}.  In this application, the attacker modifies training data either to degrade accuracy overall or to promote a target item at test time. Matrix factorization based recommender systems are vulnerable to such attacks in addition to attacks wherein the poison data is designed to appear legitimate \citepp{li2016data}. Matrix factorization methods can also be exploited by computing maximally damaging combinations of product ratings using integer programming and then deploying these using fake users \citepp{fang2020influence}. When systems select products based on user preferences and product properties, fake users can similarly inject malicious product ratings to cause promotion of target items to real customers  \citepp{fang2018poisoning}. In social networking settings, poisoning attacks can artificially promote individuals or groups as recommended connections for users \citepp{hu2019targeted}. 

\textbf{Differential privacy.} \textit{Differentially private} (DP) training is a framework for learning from data without relying heavily on or memorizing individual data samples. While originally developed to preserve user privacy, DP training also conveys a degree of resistance to data poisoning since modifications to a small number of samples can have only a limited impact on the resulting model \citepp{ma2019dplearners}. Bounds on the poisoning vulnerability of recommendation systems that use DP matrix factorization techniques have been derived and tested empirically \citepp{wadhwa2020data}. 

At the same time, differential privacy can also enable poisoning, as attackers can mask their behavior by manipulating data and model updates before DP data aggregation is applied to harvest data on a privacy-preserving central server \citepp{cao2019data}. 
Similarly, crowd sensing systems collect sensory data from people who are carrying sensor rich hardware like smartphones. These systems often employ techniques to discern which workers are contributing truthful data, as aggregation like this is prone to malicious activity. Even with truth discovery as a standard component of these systems, two poisoning attacks have been developed to harm the integrity of crowd sensing systems \citepp{miao2018towards}.

\textbf{Reinforcement learning.} Reinforcement learning algorithms are also susceptible to poisoning attacks \citepp{ma2018data}{liu2019data}. Contextual bandits, often used in adaptive medical treatment, can be manipulated by malicious changes to the rewards in the data \citepp{ma2018data}. Online learning algorithms can also be poisoned. For example, an attacker targeting stochastic bandits may perturb the reward after a decision with the goal of convincing the agent to pull a suboptimal arm \citepp{liu2019data}. Other online learning algorithms have been shown to be vulnerable to an attack that is formulated as a stochastic optimal control problem  \citepp{zhang2020online}. In this setting, an attacker can intercept a data stream providing sequential data to an online learner and craft perturbations by adopting either a model-based planning algorithm or a deep reinforcement learning approach \citepp{zhang2020online}.

\textbf{Facial recognition.} Social media users often post public images, which are in turn scraped by private or governmental organizations for unauthorized purposes including facial recognition. A number of authors have proposed poisoning face images on social media sites to prevent their use in facial recognition systems. Fawkes uses feature collisions to create images that cannot be matched to their clean counterparts by a classifier trained on small image datasets \citepp{shan2020fawkes}, but this method was subsequently found to be ineffective against real-world systems.  In a concurrent work, FoggySight instead requires users to upload decoy photos to protect others \citepp{evtimov2020foggysight}.  The LowKey system \citepp{cherepanova2021lowkey} uses large datasets, data augmentation, and an ensemble of surrogate models to create perturbations that evade matching by state-of-the-art and commercial identification systems.  The LowKey system has additionally been released as a web tool.  A similar approach, Face-Off, has recently been used to attack commercial APIs, and this work also considers the issue of adversarial defense against these attacks \citepp{gao2020face}.  

\textbf{Federated learning.}  
The distributed nature of federated learning raises a number of unique issues in dataset security.  Because model updates are harvested from a diffuse network of untrusted users, very strong adversarial threat models are realistic.  For example, it is reasonable to assume that an adversary has complete control over input data to the system, labels on training data, and model updates broadcast to the central server \citepp{bhagoji2019analyzing}{sun2020data}.  At the same time, federated learning adversaries are weak in that they only have access to a small slice of the training data -- the entire training system comprises many users that sample from diverse distributions unknown to the attacker  \citepp{tolpegin2020data}. 

Finally, user privacy is a major issue in federated learning, and privacy may be at odds with security. {\em Secure aggregation} uses cryptographic methods to ``mask''  each user's update before it is sent to the central server, making it impossible to screen incoming models for corruptions \citepp{bhagoji2019analyzing}.  Similar issues with differential privacy are discussed above.

\subsection{Feature Collision Attacks}
\label{sec:feature-collision}
 
 Feature collision attacks operate in the targeted data poisoning setting, in which the objective is to perturb training data so that a particular \emph{target example}, $x_t$, from the test set is misclassified into the \emph{base class}.  Feature collision methods for targeted data poisoning perturb training images from the base class so that their feature-space representations move towards that of the target example.  Intuitively, the attacker hopes that by saturating the region of feature space surrounding the target example with samples from the base class, the learning algorithm will classify this region into the base class, thus misclassifying the target example.  We begin with description of the original attack from \citett{shafahi2018poison}, and we then explore a variety of other feature collision attacks which have emerged since this work.
 
Following the notation of \citett{schwarzschild2020just}, let perturbed poison examples be denoted by $X_p = \{x_p^{(j)}\}_{j=1}^J$ and the corresponding corresponding samples from the original training dataset $X_b = \{x_b^{(j)}\}_{j=1}^J$, where the latter are taken from the base class.  The collision attack fixes a pre-trained feature extractor, $f$, and solves the following optimization problem:
\begin{equation}
    x_p^{(j)} = \argmin_{x} \Vert f(x) - f(x_t) \Vert_2^2 + \beta\Vert x - x_b^{(j)} \Vert_2^2.
\end{equation}
The first term in this loss function encourages the feature vector extracted from the poison example to lie close to that of the target, while the second term encourages the poison to stay close to the corresponding original sample in pixel space.  This second term promotes the \emph{clean-label} property whereby the poisoned images look like their respective base images, and thus appear to be labeled correctly.  A variant of this attack, BlackCard \citepp{guopractical}, adds two additional loss terms that explicitly encourage the poison image to be given the same label as the base image while lying far away from the base image in feature space.  These modifications lead to superior transferability to black-box models.
 
A different style of feature collision attacks aims instead to surround the target by poisons in feature space so that the feature vectors corresponding to poison examples are the vertices of a convex polytope containing the target’s feature vector~\citepp{zhu2019transferable}.  These attacks anticipate that the whole region inside the convex polytope will be classified as the base class, resulting in better attack reliability compared to a simple feature collision attack.
The creators of the Bullseye Polytope attack \citepp{aghakhani2020bullseye} notice that the target image often lies far away from the center of the polytope, leading to failed attacks.  They optimize the vertices of the polytope with the constraint that the target image is the mean of the poison feature vectors, resulting in boosted reliability.
Both polytope-based methods compute their attack on an ensemble of models to achieve better transferability in the black-box setting.  Nonetheless, in \citett{schwarzschild2020just}, feature collision methods are
shown to be brittle in the black-box setting when the victim’s architecture and training hyperparameters are unknown.
 
Feature collision poisoning methods are well-suited for the transfer learning setting, in which a model is pre-trained on clean data and fine-tuned on a smaller poisoned dataset.  This may be done by freezing the feature extractor and only fine-tuning a linear classification head.  In this setting, the attacker crafts poisons on their own surrogate models and anticipates that the feature representations of the victim model will be similar to those of their surrogate.  This threat model is quite realistic in the setting of {\em model-reuse attacks}, which exploit the fact that most transfer learning is done from standard public reference models (e.g. the pre-trained ImageNet models that ship with standard libraries).  An attacker can break many transfer learned systems by collecting a large battery of standard models and creating poisons that cause feature collisions for all of them \citepp{ji2018model}.  In Section \ref{sec:backdoor}, we also discuss a second style of transfer learning attack in which pre-training data, rather than fine-tuning data, is poisoned so that even when a model is fine-tuned on clean data, poisoning persists.  

\subsection{Bilevel Optimization}
While data poisoning in general can be formalized as a bilevel optimization problem, in this section, we discuss methods which directly approach the bilevel problem. These methods work by simulating a training pipeline, and then optimizing through this pipeline to directly search for poison data that result in corrupted models. While feature collision attacks are most effective when deployed against transfer learning, bilevel methods are highly effective against both transfer learning and end-to-end training.  Simple bilevel formulations rely on a problem of the form
\begin{equation}
    \begin{array}{rl}
     \min_{X_p  \in \mathcal{C}}& \mathcal{L} ( F(x^t, \theta'), y^{adv}),\\
     &\\
  \st&\theta' = \argmin_\theta \mathcal{L}(F(X_p \cup X_c, \theta), Y),
    \end{array}
\end{equation}
where $\mathcal{C}$ denotes a set of permissible poisons (for example within an $\ell_{\infty}$ ball around clean training data), and $F$ denotes a neural network with parameters $\theta$. In words, this formulation searches for poison images with the property that, after training on both poisoned and clean images to obtain parameters $\theta'$, the resulting model places the target image into the desired class. Some works in this space also investigate additional threat models such as untargeted attacks in which the adversary seeks to maximize average test loss rather than targeted misclassification on a single test sample \citepp{munoz2017towards}{biggio2012poisoning}.  Fundamentally, data poisoning can be formulated as a bilevel problem, and methods in the other sections of our survey can be seen as heuristic approaches to this problem.  In this section, we discuss methods which directly tackle the bilevel problem.

Early works on data poisoning for classical models solve a bilevel optimization problem, but they require that the inner problem can be solved exactly \citepp{biggio2012poisoning}{xiao2015feature}{ mei2015using}{koh2017influencefunction}.  \citett{biggio2012poisoning} use this approach to induce general performance degradation in support vector machines.  Similar algorithms have poisoned LASSO, ridge regression, and the elastic net for feature selection in malware detection \citepp{xiao2015feature}.  Another work poisons SVM on data streams \citepp{burkard2017analysis}, and is the first to study targeted data poisoning rather than general performance degradation.  \citett{mei2015using} prove that the bilevel optimization approach yields optimal training set attacks on simple models.  \citett{jagielski2018manipulating} further improve the performance of bilevel optimization based attacks on regression by strategically selecting which data to poison and also manipulating response variables.  Their work additionally provides an improved defense with formal guarantees.

For neural networks and other non-convex problems, bilevel optimization is more complex.  \citett{munoz2017towards} perform bilevel optimization with neural networks using a method they call ``back-gradient descent’’ in which the inner problem is approximately solved using several steps of gradient descent.  A gradient descent step is then conducted on the outer loss by back-propagating through the inner minimization routine.  Differentiating through multiple steps of the inner SGD is memory intensive, and so one poison sample is crafted at time instead of jointly optimizing poisons simultaneously.  Additionally, in \citett{munoz2017towards}, the back-gradient method is only applied to a single-layer network.

A recent method, MetaPoison \citepp{huang2020metapoison}, scales up this unrolling approach to bilevel optimization using realistic network architectures and training processes.  MetaPoison employs an ensembling method that uses models pre-trained with various numbers of epochs so that poison images are more likely to influence models during all parts of training.  Additionally, MetaPoison crafts all poisons simultaneously and uses adversarial perturbations to the color mapping of an image to preserve the clean label property.  This work emphasizes the transferability of poisons in the black-box setting, and successfully poison an industrial API, Google Cloud AutoML.  MetaPoison achieves significantly better performance than earlier methods, including feature collision techniques, in both the fine-tuning and from-scratch regimes on modern neural networks. The large scale ensembling is expensive though, requiring numerous GPUs to run efficiently \citepp{huang2020metapoison}.
 
Witches’ Brew \citepp{geiping2020witches} improves on MetaPoison by introducing a ``gradient alignment'' objective that encourages the gradient of the loss on poison data to match the gradient of the adversary's target loss.  When these two gradients are aligned, standard gradient descent steps for training on the poisoned images will also decrease the adversarial loss on the target images, causing model poisoning.  Improved computational efficiency enables this work to conduct the first targeted data poisoning of from-scratch training on ImageNet.  The method was also demonstrated to break the Google Cloud AutoML API on ImageNet training. \citett{schwarzschild2020just} benchmark both the transfer learning and from-scratch settings of targeted data poisoning, and they find that while Witches' Brew is the highest performing method on the from-scratch CIFAR-10 \citepp{krizhevsky2009learning} problem, Bullseye Polytope actually outperforms this method on the higher dimensional Tiny-ImageNet dataset \citepp{le2015tiny}.

Another work, focusing on secure dataset release, harnesses the Witches' Brew gradient matching objective to poison every image in a dataset and render the models trained on the poisoned dataset useless for achieving high test accuracy \citepp{fowl2021preventing}.  This method can also be employed effectively in an online fashion in which social media companies upload perturbed user images to protect their proprietary data from competitors who may attempt to leverage the data to improve their own models.
 
Solving the bilevel optimization problem is computationally expensive, even using Witches’ Brew.  Several methods thus propose to bypass this problem by training generative models such as GANs and autoencoders to produce poisons that mimic those generated by bilevel methods \citepp{yang2017generative}{munoz2019poisoning}.  After a generative model is trained, poisons can be crafted simply by conducting a forward pass through the generator or the autoencoder.  While this enables poisoning additional images at little additional computational cost, these methods may be less effective, and the cost of training the original generative model is high.

Instead of aligning the gradient of training data with a targeted misclassification loss, TensorClog \citepp{shen2019tensorclog} poisons data to cause a vanishing gradient problem during training.  This method perturbs training data so that neural networks have gradients of very low magnitude and thus do not train effectively while also ensuring that perturbed images have a high value of SSIM with respect to the originals.  TensorClog both prevents networks from achieving low loss during training while also decreasing validation performance.  However, this strategy is only effective in the white-box transfer learning setting when the feature extractor is known and fixed, and even in this idealized setting, the attack is only mildly successful \citepp{shen2019tensorclog}.  A similar method minimizes training loss w.r.t. poisoned images in order to prevent data from being useful for training networks \citepp{huang2021unlearnable}.  A generative model based method instead uses autoencoders to generate perturbations which yield poor overall test accuracy in models which train on perturbed training data \citepp{feng2019learning}.  This method approaches the same gradient matching objective as \citett{fowl2021preventing} opting for a generative model instead of solving the bilevel problem directly.

\subsection{Label Flipping}
Label flipping attacks opt to switch training labels while leaving the data instances untouched.  While these attacks are not ``clean-label'', they have the advantage of not introducing strange-looking artifacts, which may be obvious to the intended victim. \citett{biggio2011support} use both random and adversarial label flips to poison support vector machines.  Their work shows that flipping the labels of an adversarially chosen data subset can cause poisoning, even against learners trained in a robust fashion. \citett{zhao2017efficient} show that a projected gradient ascent approach to label flipping is effective even against black-box linear models ranging from SVM to logistic regression and LS-SVM.  \citett{zhang2017game} provide a theoretical analysis of label flipping attacks on SVM using tools from game theory.  In response to these attacks, a number of defenses, theoretical and empirical, have emerged against label flipping \citepp{paudice2018label}{rosenfeld2020certified}.  In regression, response variables can be manipulated to enhance data-perturbation based poisoning \citepp{jagielski2018manipulating}.

\subsection{Influence Functions}
Influence functions estimate the effect of an infinitesimal change to training data on the model parameters that result from training, which can be leveraged to construct poisoning instances.  This measurement can then can be employed to approximate solutions to the bilevel formulation of poisoning \citepp{koh2018stronger}.  A function for measuring the impact of up-weighting a data sample can be written as
\begin{equation}
\begin{split}
    \mathcal{I}(x) = -H_{\theta'}^{-1}\nabla_{\theta}\mathcal{L}(f_\theta'(x)),\\
    \theta' = \argmin_{\theta} \sum_{i=1}^{n}\mathcal{L}(f_{\theta}(x_i)),
\end{split}
\end{equation}
where $H_{\theta'}$ denotes the Hessian of the loss with parameters $\theta'.$
This influence function can be leveraged to compute the influence that removing a particular data point would have on test loss \citepp{koh2017influencefunction}.  This approach can indicate how individual training samples are responsible for specific inference-time predictions on linear models whose input features are extracted by a neural network. \citett{koh2017influencefunction} also extend the influence function method to create ``adversarial training examples’’ as well as to correct mislabelled data \citepp{koh2017influencefunction}.  \citett{koh2018stronger} leverage such influence functions to create stronger poisoning attacks by accelerating bilevel optimization.  \citett{fang2020influence} adopt this approach to conduct data poisoning attacks on recommender systems.
It is worth noting that, while influence functions have been successful on classical machine learning algorithms, they do not effectively capture data dependence in modern deep neural networks which have highly non-convex loss surfaces \citepp{basu2020influence}.

\subsection{Online Poisoning}
The concept of online poisoning originated with the study of data corruption where each bit is perturbed with probability $p$~\citepp{valiant1985learning}{kearns1993learning}.
When these attacks are restricted to be clean-label, they are known as \emph{$p$-tampering} attacks and were originally studied in the context of security~\citepp{austrin2014impossibility}. \citett{mahloujifar2017blockwise} consider the setting of targeted poisoning and expand the idea to block-wise $p$-tampering, where entire samples in the training set of a learning pipeline can be modified by an adversary. This work studies poisoning attacks from a theoretical standpoint and includes several results concerning the vulnerability of a model given the portion $p$ of the data that is perturbed. Specifically, Mahloujifar and Mahmoody prove the existence of effective targeted clean label attacks for deterministic learners. \citett{mahloujifar2018learning} further improve the quantitative bounds. Moreover, they extend $p$-tampering to the untargeted case in the PAC learning setting, where they prove the existence of attacks that degrade confidence.  Further theoretical work relates $p$-tampering attacks to concentration of measure, wherein the existence of poisoning attacks against learners over metric spaces is proven \citepp{mahloujifar2019curse}.

\subsection{Data Poisoning in Federated Learning}
\label{sec:poisoning-fl}

In the federated learning setting, an adversary can insert poisons at various stages of the training pipeline, and attacks are not constrained to be ``clean-label’’ since the victim cannot see the attacker’s training data.  \citett{tolpegin2020data} study targeted label flipping attacks on federated learning.  Their work finds that poisons injected late in the training process are significantly more effective than those injected early.  Another work instead adopts a bilevel optimization approach to poisoning multi-task federated learning \citepp{sun2020data}.  Others instead opt for GAN-generated poisons \citepp{zhang2019poisoning}.
 
\emph{Model poisoning} is a unique poisoning strategy for federated learning; in contrast to data poisoning, the adversary directly manipulates their local model or gradient updates without the need to modify data or labels.  One such approach performs targeted model poisoning in which the adversary ``boosts'' their updates to have a large impact on the global model while still staying beneath the radar of detection algorithms \citepp{bhagoji2019analyzing}.  The approach is effective even against Byzantine-resilient aggregation strategies.  Another approach shows that Byzantine-robust federated learning methods can be broken by directly manipulating local model parameters in a ``partial knowledge’’ setting in which the attacker has no knowledge of local parameters of benign worker devices \citepp{fang2020local}.
 
\emph{Sybils} are groups of colluding agents that launch an attack together.  \citett{cao2019understanding} develop a distributed label flipping attack for federated learning and study the relationship between the number of sybils and the attack’s effectiveness.  \citett{fung2018mitigating} also study label flipping based sybil attacks, as well as backdoors, for federated learning.  Their work additionally proposes a defense called ``FoolsGold’’ that de-emphasizes groups of clients who's contributions to the model are highly similar.

Theoretical work has studied $p$-Tampering for multi-party learning, of which federated learning is a special case \citepp{mahloujifar2019data}.  However, this work assumes that the adversary has knowledge of updates generated by other benign parties.

\subsection{Open Problems}
\begin{itemize}
\item \textbf{Accelerated data poisoning for from-scratch training:} While bilevel optimization based data poisoning algorithms are significantly more effective than feature collision at poisoning neural networks trained from scratch, they are also computationally expensive.  Developing effective data poisoning algorithms for industrial-scale problems is a significant hurdle.
\item \textbf{Attacking with limited information about the dataset:}  Existing training-only attacks implicitly assume the attacker knows the whole dataset being used for training.  More realistic settings would involve an attack with limited knowledge of the dataset, or even the exact task, being solved by the victim.
\item \textbf{True clean-label attacks:}  Existing works often permit large input perturbation budgets, resulting in poison images that are visibly corrupted.  The adversarial attack literature has recently produced a number of methods for crafting fairly invisible adversarial examples. Adapting these methods to a data poisoning setting is a promising avenue towards truly clean-label attacks.
\item \textbf{Fair comparison of methods:}  Experimental settings vary greatly across studies.  One recent benchmark compares a number of attacks across standardized settings \citepp{schwarzschild2020just}.  Nonetheless, many methods still have not been benchmarked, and a variety of training-only threat models have not been compared to the state of the art.
\item \textbf{Robustness to the victim’s training hyperparameters:} \citett{schwarzschild2020just} also show that many existing poisoning methods are less effective than advertised, even in the white-box setting, when attacking network architectures, optimizers, or data augmentation strategies that differ from the original experimental setting.  Ongoing work seeks attacks that transfer to a wide range of training hyperparameters.  The success of MetaPoison \citepp{huang2020metapoison} and Witches’ Brew \citepp{geiping2020witches} at poisoning AutoML suggests that this goal is achievable, but quantifying robustness across hyperparameters in a controlled testing environment remains a challenge. 
\item \textbf{Broader objectives:}  Most work on training-only data poisoning has focused on forcing misclassification of a particular target image or of all data simultaneously. However, broader goals are largely unexplored.
For instance, one can aim to cause misclassification of an entire sub-class of inputs by targeting specific demographics~\citepp{jagielski2020subpopulation} or inputs corresponding to a particular physical object or user.  
\end{itemize}

\section{Backdoor Attacks}
\label{sec:backdoor}

\begin{figure*}[h!]
    \centering
    \resizebox{\textwidth}{!}{
        \begin{tikzpicture}[
            i/.style={
                anchor=north,
                align=center,
                top color=white,
                bottom color=blue!10,
                rectangle,rounded corners,
                minimum height=6mm,
                draw=blue!75,
                align=center,
                color=blue!75,
                text=black,
                text depth = 0pt
            }
        ]
            \tikzset{execute at begin node=\strut}
            \tikzset{font=\small,
                grow=down,
                level distance=1.5cm,
                every tree node/.style={align=center,anchor=north},
                every leaf node/.style=
                    {
                        anchor=north,
                        align=center,
                        fill=blue!5,
                        rectangle,
                        draw=blue!50,
                        align=center,
                        distance=1cm,
                        text depth = -10pt
                    },
                edge from parent/.style=
                    {
                        draw=blue!50,
                        thick,
                        edge from parent path={(\tikzparentnode.south)
                        -- +(0,-8pt)
                        -| (\tikzchildnode)}
                    }
            }

            \Tree [.\node[i] {Backdoor Attacks};
            [
                .\node[i] {Applications};
                [
                    .\node[i] {Object Recognition \\ and Detection\\
                    \cite{chen2017targeted},
                            \cite{gu2017badnets}};
                ]
                [
                     .\node[i] {Generative \\ Models\\
                     \cite{ding2019trojan},
                        \cite{salem2020baaan},
                        \cite{schuster2020you},\\ \cite{wallace2020customizing},
                        \cite{zhang2020trojaning}};
                ]
                [
                    .\node[i] {Reinforcement \\ Learning\\
                    \cite{kiourti2020trojdrl},
                         \cite{wang2020stop},
                         \cite{yang2019design}};
                ]
                [
                    .\node[i] {Model \\ Watermarking\\
                    \cite{adi2018turning},
                         \cite{zhang2018protecting}};
                ]
            ]
            [
                .\node[i] {Methodology};
                [
                    .\node[i] {Basic \\ Attacks\\
                    \cite{chen2017targeted},
                            \cite{gu2017badnets}};
                ]
                [
                    .\node[i] {Backdooring \\ Pre-trained Models\\
                    \cite{liu2017trojaning},
                        \cite{sun2020poisoned},
                        \cite{tang2020embarrassingly}};
                ]
                [
                    .\node[i] {Clean-label \\ Attacks\\
                    \cite{saha2019hidden} , 
                        \cite{turner2019label},
                        \cite{wallace2020customizing}};
                ]
                [
                    .\node[i] {Attacks for \\ Transfer Learning\\
                    \cite{wang2020backdoor},
                        \cite{yao2019latent}};
                ]
                [
                    .\node[i] {Attacks for \\ Federated Learning\\
                    \cite{bagdasaryan2020backdoor},
                         \cite{baruch2019little},
                         \cite{chen2020backdoor},\\
                         \cite{liu2020backdoor},
                         \cite{sun2019can},
                         \cite{xie2019dba} };
                ]
            ]
            ]
        \end{tikzpicture}
    }
    \caption{A taxonomy of backdoor attacks.}
    \label{fig:backdoor-tree}
\end{figure*}
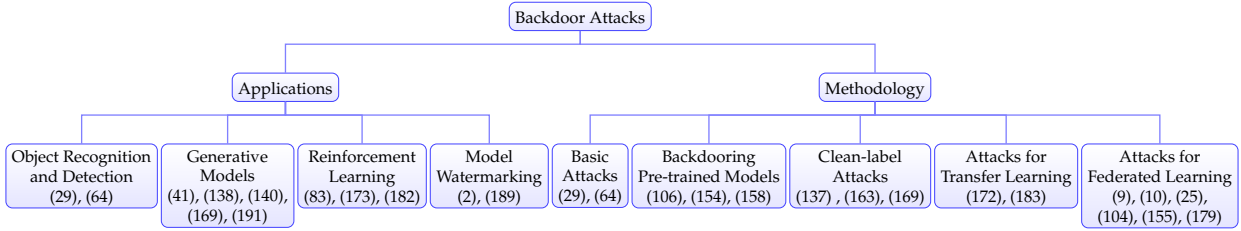

In contrast to the attacks discussed above, backdoor attacks (also known as Trojan attacks) allow the adversary (limited) access to inputs during inference.
This capability allows the adversary to perform significantly more potent attacks, changing the behavior of the model on a much broader range of test inputs.  Figure \ref{fig:backdoor-tree} contains a taxonomy of backdoor attacks according to different objectives and threat models.

\begin{figure}
    \vspace{-2mm}
    \centering
    \hfill
    \begin{subfigure}[t]{0.3\linewidth}\centering
    \includegraphics[width=\linewidth]{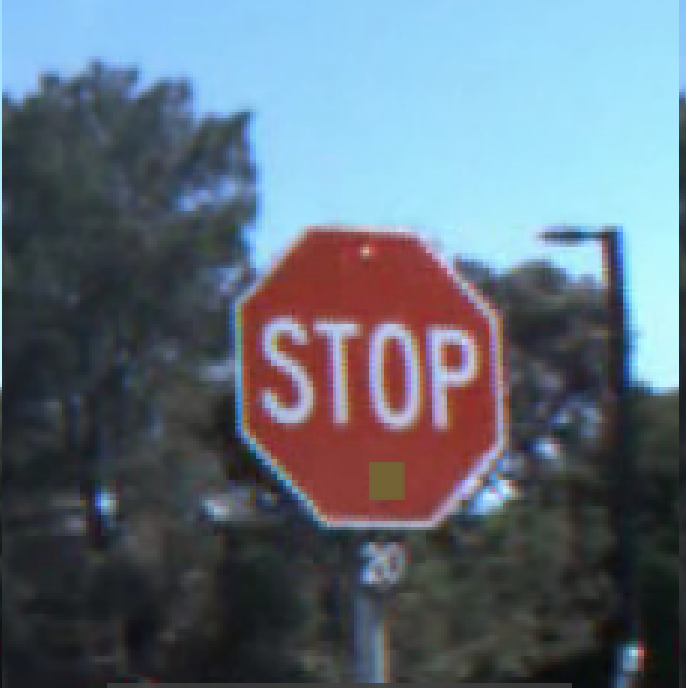}
    \caption{}
    \label{fig:backdoor-trigger-square}
    \end{subfigure}
    \hfill
    \begin{subfigure}[t]{0.45\linewidth}\centering
    \includegraphics[width=\linewidth]{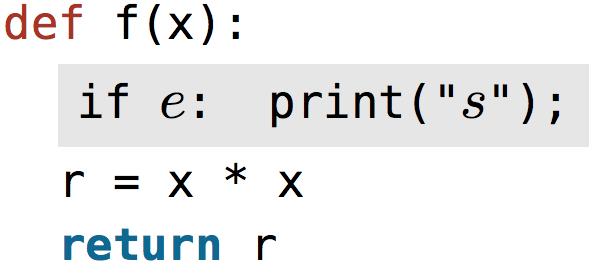}
    \caption{}
    \label{fig:backdoor-trigger-code}
    \end{subfigure}
    \hfill
    \phantom{}
    \caption{Backdoor attacks with different triggers: (a) a square pattern flips the identity of a stop-sign (\citeauthor{gu2017badnets} \citeyear{gu2017badnets}); 
    (b) dead code as the trigger for source code modeling (\citeauthor{ramakrishnan2020backdoors} \citeyear{ramakrishnan2020backdoors}).}
    \vspace{-2mm}
    \label{fig:backdoor-trigger}
\end{figure}

The key idea behind this class of attacks is to poison a model so that the presence of a \emph{backdoor trigger} in a test-time input elicits a particular model behavior (e.g. a particular label assignment) of the adversary's choice.
The trigger is a pattern that is easily applied to any input --- e.g., a small
patch or sticker in the case of images or a specific phrase in the case of natural language processing (cf. Figure~\ref{fig:backdoor-trigger}).
To ensure that the attack goes undetected when the model is actually deployed, it is necessary that the model behaves normally in the absence of a trigger, i.e., during normal testing.


The most common scenario for backdoor attacks involves end-to-end training~\citepp{gu2017badnets}{chen2017targeted}, where the adversary is able to inject multiple poisoned inputs into the training set, causing a backdoor vulnerability in the trained model.
This scenario is quite general and encompasses several real-world ML tasks---e.g., text classification \citepp{dai2019backdoor}{chen2020badnl}{sun2020natural}, graph classification \citepp{zhang2020backdoor}{xi2020graph}, malware detection \citepp{severi2020exploring}, biometric systems \citepp{lovisotto2019biometric}, and reinforcement learning~\citepp{kiourti2020trojdrl}.

Another threat model corresponds to the setting where the model is not trained from scratch but rather fine-tuned on a new task.
Just as in the case of training-only attacks,
the adversary can exploit access to commonly used standard models to produce more potent attacks.


\paragraph{Comparison to evasion attacks.} At a high level, backdoor attacks appear similar to evasion attacks \citepp{biggio2013evasion}{szegedy2013intriguing} where the adversary perturbs a specific input to cause an intended model prediction.
The key difference here is that backdoor attacks aim to embed a trigger that is \emph{input- and model-agnostic}---i.e., the same trigger can cause any poisoned model to produce an incorrect prediction on any input.
While in principle it is possible to construct evasion attacks that apply to multiple inputs \citepp{moosavi2017universal} and transfer across models \citepp{szegedy2013intriguing}, such attacks are less effective \citepp{tramer2017space}.

\subsection{Applications of Backdoor Attacks}
\label{sec:backdoor-applications}
Backdoor attacks can be used to manipulate learning systems in a range of application areas that we survey below.


\paragraph{Object recognition and detection.} 
Early work on backdoor attacks focused on manipulating computer vision systems by altering physical objects. 
\citett{gu2017badnets} show that their backdoor attack can induce an image classifier to label a stop sign as a speed limit sign in the presence of a small yellow sticker that acts as the trigger.
Similarly, \citett{chen2017targeted} demonstrate backdoor attacks on a simplified face recognition system in which a pair of glasses triggers a prediction change in identity.

\textbf{Generative models.}
The danger posed by backdoor attacks is not restricted to classifiers; they can be applied to models whose output is not a single label.
In the case of language models, a trigger can elicit complex behaviors such as generating specified character sequences.  \citett{zhang2020trojaning} use a trigger phrase to cause generative language models to produce offensive text completion. Similar results occur in machine translation \citepp{wallace2020customizing}, when the trigger phrase appears in the context of the phrase being translated.
Other objectives include suggesting insecure source code \citepp{schuster2020you} or generating images with specific characteristics \citepp{ding2019trojan}{salem2020baaan}.

Adapting attacks on classification to work on generative models is not always straightforward.
One challenging aspect is that the poisoned inputs may need to obey a range of application-specific constraints.
For instance, backdoor attacks on natural language systems \citepp{dai2019backdoor} may require the poisoned inputs to be natural and syntactically valid.
To achieve this, $\text{TROJAN}^{\text{LM}}$ \citepp{zhang2020trojaning} fine-tunes a pre-trained GPT-2 model~\citepp{radford2019language} to generate sentences containing specified keywords when triggered.
In source code modeling, the trigger may need to be injectable without causing runtime errors or behavioral changes, which can be achieved by only modifying ``dead'' code paths that can never be executed \citepp{ramakrishnan2020backdoors}.

\textbf{Reinforcement learning.}
Backdoor attacks on reinforcement learning aim to cause the agent to perform a malicious action when the trigger appears in a specific state (e.g., a symbol on the screen of an Atari game)~\citepp{yang2019design}{kiourti2020trojdrl}.
For example, in traffic systems, the attacker's goal may be to cause congestion when a specific traffic pattern is observed \citepp{wang2020stop}. One key challenge is that, for the attack to remain unnoticed, the trigger should be added to as few states as possible.
Thus, from the attacker's perspective, it is desirable that the adversarial behavior persists even after the trigger disappears from the agent's observation \citepp{yang2019design}.

\textbf{Model watermarking.} Backdoor attacks rely on the expressiveness of machine learning models and in particular DNNs---being able to memorize specific patterns without degrading their overall accuracy.
This property of DNNs has also been leveraged for model watermarking \citepp{adi2018turning}{zhang2018protecting}. The goal of watermarking is to train a model while ensuring that a predetermined set of patterns or inputs gets assigned specific (randomly chosen) labels by the model. 
Then, should an adversary steal and deploy the model for profit, the model owner can prove ownership by demonstrating knowledge of the embedded watermarks---e.g., using standard cryptographic primitives \citepp{adi2018turning}.

\subsection{Basic Backdoor Attacks}
The most common paradigm for launching backdoor attacks is to inject the dataset with poison samples containing the backdoor trigger. A desirable property of such attacks is that they are \emph{model-agnostic} so that the same backdoor attack is effective on different models trained on the same poisoned dataset. Therefore, the attacks can be launched in the \emph{black-box} scenario.

The first successful backdoor attacks on modern deep neural networks are demonstrated in \citett{gu2017badnets} and \citett{chen2017targeted}, where the adversary injects mislabelled poison samples into the training set to perform the attack. To encourage the model to rely on the trigger, the adversary chooses a number of natural images and labels them incorrectly with the target class label before adding the backdoor trigger.
The resulting images are therefore mislabelled based on their content and are only associated with their label via the backdoor trigger.
During training, the model strongly relies on the easy-to-learn backdoor trigger in order to classify these images.
As a result, when the trigger is applied to a new image during deployment, the model assigns it the target label, as desired by the adversary.

\citett{gu2017badnets} demonstrate their attack in the setting where the model is being trained by the adversary, but \citett{chen2017targeted} show that the method also works without access to model training and with significantly fewer poison examples.
Moreover, these attacks have been shown to be effective with imperceptible triggers in the training set \citepp{chen2017targeted}{li2020invisible} while still being realizable in the physical world \citepp{gu2017badnets}{chen2017targeted}{wenger2020backdoor}{sarkar2020facehack}.  \citett{dai2019backdoor} similarly attack RNNs for text classification.  They show that simple and unnoticeable phrase modifications can reliably induce backdoor behavior.

\subsection{Model-Specific Attacks}
\label{sec:backdoor-model-specific}
The backdoor attacks described so far have the property that they are model-agnostic---i.e., the triggers are simple patterns constructed without any knowledge of the model under attack.
In this section, we describe how one can mount more powerful attacks by tailoring the attack to a particular model.

\subsubsection{Embedding Backdoors into Pre-Trained Models} In the case where the attacker obtains access to an already trained model,
they can embed a backdoor into this model without completely re-training it and even without access to the original training data. 
\citett{liu2017trojaning} select an input region along with neurons which are sensitive to changes in that region, and the attacker aims to activate these neurons. They then generate artificial data (from the model alone), and they adapt the basic backdoor attack by applying their trigger to this artificial data and fine-tuning only a few layers of the model. \citett{tang2020embarrassingly} propose an approach that adds a small module to the existing model instead of modifying existing model weights. Note that this attack does modify the model architecture and hence falls outside the standard threat model. \citett{sun2020poisoned} start from a model into which a backdoor has already been injected and focus on constructing alternative backdoor triggers. They find that it is possible to construct multiple distinct triggers without knowledge of the original one.

\subsubsection{Clean-Label Backdoor Attacks} 
Most work on backdoor attacks requires adding mislabeled poison samples into the training set.
However, such attacks assume that the adversary has access to the labeling process. Moreover, they are likely to be detected should a human manually inspect these samples.
Recent work proposes clean-label backdoor attacks, where the labels of poison samples aim to be semantically correct \citepp{turner2019label}{saha2019hidden}.

The methodology behind clean-label backdoor attacks is conceptually similar to the feature collision methods introduced in Section~\ref{sec:feature-collision}.
\citett{turner2019label} utilize generative adversarial networks \citepp{goodfellow2014generative} and adversarial examples \citepp{szegedy2013intriguing} to perturb images of the target class towards other classes (hence making them harder to learn) before applying the backdoor trigger. 
The Hidden Trigger Backdoor Attack \citepp{saha2019hidden} adapts the feature collision framework \citepp{shafahi2018poison} to construct backdoor poison samples.
These samples are based off natural images from the target class but slightly modified so that their feature representations are close to images injected with the backdoor trigger.

For natural language processing, \emph{no-overlap} backdoor attacks are designed to make the poison samples hard to identify \citepp{wallace2020customizing}. Specifically, the input sentences of poison samples in the training set do not contain the words in the backdoor trigger, yet when the test-time input contains the backdoor trigger, a poisoned model still produces the adversarial prediction. The poison samples are generated by manipulating the model gradient during training, similar to the gradient-based optimization procedure in \citett{munoz2017towards}.

\subsubsection{Backdoor Attacks for Transfer Learning}
So far, we have discussed backdoor attacks where the victim either trains the model from scratch on a poisoned dataset or receives an already trained model.
A scenario that interpolates between these two settings is transfer learning, in which part (or all) of the model is re-trained on a new task.
In contrast to the transfer learning setting from Section \ref{sec:train} where fine-tuning data was poisoned, the threat model we discuss in this section consists of an adversary that poisons the pre-training data to create a backdoored feature extractor, but has no control over the victim's fine-tuning process.
Basic backdoor attacks can still be harmful in this setting when only the final fully connected layer of the model is re-trained \citepp{gu2017badnets}.
However, when the entire model is fine-tuned in an end-to-end fashion, the embedded backdoor is virtually eliminated \citepp{liu2018fine}{liu2017neural}.

To ensure that the backdoor trigger is persistent after fine-tuning, existing works design trigger patterns specific to the attacker's goal \citepp{yao2019latent}{wang2020backdoor}. To generate the optimal backdoor trigger for each target label, the adversary annotates a number of clean samples with the target label, injects them into the training set, and trains the model on this augmented dataset. Then, the adversary generates the backdoor trigger that maximizes the activation of neurons responsible for recognizing the target label, and applies the basic backdoor attack to generate poison samples for further model pre-training. In order to generate the final trigger, the adversary optimizes the color intensity so that the intermediate feature representations of inputs injected with the trigger are close to clean samples in the target class. In \citett{yao2019latent}, the authors show that when all layers after the intermediate layer for trigger generation are frozen, the attack remains effective for transfer learning, and existing backdoor defenses cannot effectively detect or eliminate the backdoor without degrading the benign prediction accuracy.

To make the attacks more resilient to pruning and fine-tuning based defenses, \citett{wang2020backdoor} propose a ranking-based neuron selection mechanism, which identifies neurons with weights that are hard to change during the pruning and fine-tuning process.  This work utilizes an autoencoder to generate strong triggers that are robust under defenses based on input preprocessing. With the proposed defense-aware fine-tuning algorithm, such backdoor attacks retain higher success rates.

\subsubsection{Backdoor Attacks on Federated Learning}
Directly applying a basic backdoor attack, i.e., training a local backdoored model and using it to update the global model, does not work in the federated setting~\citepp{bagdasaryan2020backdoor}.
The main obstacle is that aggregation across many users will reduce the effect of an individual adversarial update.
To overcome this challenge, \citett{bagdasaryan2020backdoor} study the model replacement approach, where the attacker scales a malicious model update so as to overpower other benign model updates, effectively replacing the global model with the adversary's backdoored local model.
This attack can be modified to bypass norm-based and statistics-based defenses for federated learning by constraining the norm and variance of gradient updates for local models \citepp{sun2019can}{baruch2019little}.

\citett{xie2019dba} propose distributed backdoor attacks, which better exploit the decentralized nature of federated learning. Specifically, they decompose the backdoor pattern for the global model into multiple distributed small patterns, and inject them into training sets used by several adversarial participants. Compared to injecting the global backdoor trigger, injecting separate local patterns for different participants improves the effectiveness of the attacks and bypasses robust  aggregation algorithms.

In addition to the standard federated learning scenario, where the participants have disjoint training sets for a single task, backdoor attacks have been proposed for other settings. For example, \citett{liu2020backdoor} investigate backdoor attacks for feature-partitioned federated learning, where each participant only has access to a subset of features, and most of them do not have access to labels. They demonstrate that even without manipulating the labels, the adversary can still successfully embed the backdoor, but such attacks are easier to repel with gradient aggregation mechanisms. \citett{chen2020backdoor} propose backdoor attacks for federated meta-learning, where they collaboratively train a model that is able to quickly adapt to new tasks with a few training samples. They demonstrate that the effects of such attacks still persist after meta-training and fine-tuning on benign data.

\subsection{Open Problems}

\begin{itemize}
    \item \textbf{Backdoors that persist after end-to-end fine-tuning: } While backdoor attacks remain effective when the defender freezes most layers in the pre-trained model for fine-tuning, when the entire model is fine-tuned end-to-end, existing attacks for transfer learning fail \citepp{yao2019latent}{wang2020backdoor}. More generally,  developing backdoor attacks without strong assumptions on the fine-tuning process remains a challenge for the transfer learning setting.
    \item \textbf{Backdoor attacks with limited training data:} Existing backdoor attack approaches typically require access to clean samples for training. Even if the adversary already has access to a pre-trained model, the attacks only work for specific triggers extracted from the model \citepp{liu2017trojaning}. One potential avenue for bypassing this obstacle would be to embed the trigger directly into the model weights in a method similar to existing watermarking approaches \citepp{rouhani2018deepsigns}{uchida2017embedding}.
    \item \textbf{Architecture-agnostic clean-label attacks:} The clean-label backdoor attacks described so far work best when the adversary has access to a surrogate model that closely reflects the architecture of the victim model~\citepp{turner2019label}{saha2019hidden}{wallace2020customizing}. To improve the transferability of clean-label attacks among a broad range of model architectures, one might leverage techniques for generating transferable adversarial examples~\citepp{moosavi2017universal}{liu2016delving} and clean-label attacks targeting specific instances~\citepp{huang2020metapoison}, e.g., using an ensemble of models to generate poison samples.
    \item \textbf{Understanding the effectiveness of backdoor attacks in the physical world:} Physically realizable backdoor attacks have been explored in existing works, mostly in the setting of face recognition~\citepp{chen2017targeted}{wenger2020backdoor}{sarkar2020facehack}. While these attacks can still be successful under different physical conditions, e.g., lighting and camera angles, the attack success rate drastically varies across backdoor triggers. How different factors affect the physical backdoor attacks is still an overlooked challenge, and drawing inspiration from physical adversarial examples~\citepp{eykholt2018robust}{athalye2018synthesizing} to propose robust physical backdoor attacks is another promising direction.
    \item \textbf{Combining poisoning and test-time attacks for stronger backdoor attacks:} To perform backdoor attacks at test time, the common practice is to directly embed the backdoor trigger without additional modification on the input. One potential avenue to further strengthen these attacks is to apply additional perturbations to the input aside from the trigger. The optimal perturbation could be computed in a similar way to adversarial example generation. Integrating evasion attacks with backdoor attacks could lead to improved success rates and mitigate the difficulty of backdoor embedding.
\end{itemize}

\section{Defenses Against Poisoning Attacks}
\label{sec:defenses}

\begin{figure*}[h!]
    \centering
    \resizebox{\textwidth}{!}{
        \begin{tikzpicture}[
            i/.style={
                anchor=north,
                align=center,
                top color=white,
                bottom color=blue!10,
                rectangle,rounded corners,
                minimum height=6mm,
                draw=blue!75,
                align=center,
                color=blue!75,
                text=black,
                text depth = 0pt
            }
        ]
            \tikzset{execute at begin node=\strut}
            \tikzset{font=\small,
                grow=down,
                level distance=1.5cm,
                every tree node/.style={align=center,anchor=north},
                every leaf node/.style=
                    {
                        anchor=north,
                        align=center,
                        fill=blue!5,
                        rectangle,
                        draw=blue!50,
                        align=center,
                        distance=1cm,
                        text depth = -10pt
                    },
                edge from parent/.style=
                    {
                        draw=blue!50,
                        thick,
                        edge from parent path={(\tikzparentnode.south)
                        -- +(0,-8pt)
                        -| (\tikzchildnode)}
                    }
            }

            \Tree [.\node[i] {Defenses Against Poisoning Attacks};
            [
                .\node[i] {Identifying \\ Poisoned  Data};
                [
                    .\node[i] {Outliers in \\ Input Space\\
                    \cite{diakonikolas2019sever},
                    \cite{paudice2018detection},\\
                    \cite{paudice2018label},
                    \cite{steinhardt2017certified}};
                ]
                [
                     .\node[i] {Latent \\ Space \\Signatures\\
                      \cite{chen2018activationclustering}, \cite{koh2017influencefunction},\\
                            \cite{ma2019nic}, \cite{peri2019deepknn},\\
                            \cite{tran2018spectral}};
                ]
                [
                    .\node[i] {Predictions \\Signatures\\
                    \cite{chou2020sentinet}
                         \cite{gao2019strip}};
                ]
            ]
            [
                .\node[i] {Identifying \\ Poisoned  Models};
                [
                    .\node[i] {Trigger \\ Reconstruction\\
                    \cite{ijcai2019DeepInspect},
                        \cite{guo2019tabor},\\
                        \cite{wang2019neural},
                        \cite{wang2020practical}};
                ]
                [
                    .\node[i] {Trigger-agnostic \\ Detection\\
                    \cite{huang2020one},
                        \cite{kolouri2020universal}, \\
                        \cite{xu2019Meta}};
                ]
            ]
            [
                .\node[i] {Repairing  Poisoned  Models \\ after  Training};
                [
                    .\node[i] {Patching \\ Known  \\ Triggers\\
                                            \cite{ijcai2019DeepInspect},
                        \cite{qiao2019defending},\\
                        \cite{wang2019neural},
                        \cite{zhu2020gangsweep}};
                ]
                [
                    .\node[i] {Trigger-agnostic  \\Backdoor  \\Removal\\
                    \cite{chen2019refit},
                        \cite{liu2018fine},\\
                        \cite{liu2020removing}};
                ]
            ]
            [
                .\node[i] {Preventing Poisoning \\ during  Training};
                [
                    .\node[i] {Randomized \\ Smoothing\\
                    \cite{rosenfeld2020certified},
                        \cite{weber2020rab}};
                ]
                [
                    .\node[i] {Majority \\ Vote \\ Mechanisms\\
                        \cite{jia2020certified},
                        \cite{jia2020intrinsic}, \\ \cite{levine2020deeppartition}};
                ]
                [
                    .\node[i] {Differential \\ Privacy\\
                    \cite{hong2020GradientShaping},
                        \cite{ma2019dplearners}};
                ]
                [
                    .\node[i] {Input \\ Preprocessing\\
                     \cite{borgnia2020strong}, \cite{borgnia2021dp},
                     \\  \cite{geiping2021doesn}, \cite{liu2017neural}};
                ]
            ]
            [
                .\node[i] {Defenses for \\ Federated Learning};
                [
                    .\node[i] {Robust \\ Federated \\ Aggregation\\
                                            \cite{blanchard2017machine}, 
                        \cite{chen2017distributed}, \\
                        \cite{fu2019attack}, 
                        \cite{fung2018mitigating}, \\
                        \cite{li2019rsa}, 
                        \cite{li2020learning}, \\
                        \cite{mhamdi2018hidden}, 
                        \cite{pillutla2019robust}, \\
                        \cite{yin2018byzantine}};
                ]
                [
                    .\node[i] {Robust \\ Federated \\ Training\\
                    \cite{andreina2020baffle},
                    \cite{sun2019can}};
                ]
                [
                    .\node[i] {Post-Training \\ Defenses\\
                    \cite{wu2020mitigating}};
                ]
            ]
            ]
        \end{tikzpicture}
    }
    \caption{A taxonomy of defenses against training-only and backdoor attacks.}
    \label{fig:defense-tree}
\end{figure*}
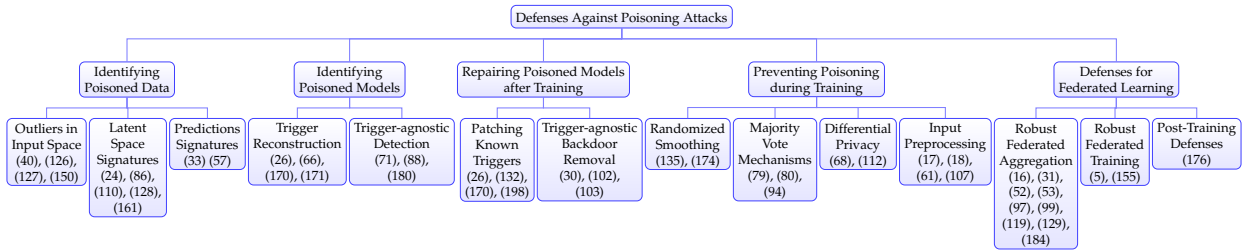

In this section, we discuss defense mechanisms for mitigating data poisoning attacks. These tools are employed at different stages of the machine learning pipeline, and can be broken down into three categories.
One type of defenses detects the existence of poisoning attacks by analyzing either the poisoned training set or the model itself.
The second class of defenses aims to repair the poisoned model by removing the backdoor behavior from the system.
The third and final group comprises robust training approaches designed to prevent poisoning from taking effect.  Figure \ref{fig:defense-tree} depicts a taxonomy of defenses against training-only and backdoor attacks according to methodology.

\subsection{Identifying Poisoned Data}
The broad goal of detection-based defense strategies is to discover axes along which poison examples or model parameters differ from their non-poisoned counterparts. These detection methods are based on raw input data or latent feature representations, or they are otherwise designed to analyze the behavior of the model around a specific input.

\subsubsection{Outliers in Input Space}
Perhaps the simplest method for detecting poisoned inputs is to identify outliers in the input space of the model.
This principle connects to a long line of work, known as \emph{robust statistics}~\citepp{huber2004robust}{hampel2011robust}, going all the way back to the work of ~\citett{tukey1960survey} and~\citett{huber1964robust}.
The high level goal of this field is to estimate statistical quantities of a dataset in the presence of (adversarially-placed) outliers.
A significant amount of work in this space shows that, from an information-theoretic point of view, this problem is indeed tractable for a wide variety of tasks and data distributions~\citepp{tukey1960survey}{donoho1988automatic}{zuo2000general}{chen2018robust}{steinhardt2018resilience}{zhu2019generalized}.
However, from a computational perspective, most of these approaches do not provide efficient implementations for high-dimensional datasets (which, after all, are at the core of modern ML).

Recently, there has been a flurry of activity focused on designing computationally efficient algorithms in this setting.
\citett{klivans2009learning} present the first algorithm for learning a linear classifier under adversarial noise. More recently, \citett{diakonikolas2019robust} and \citett{lai2016agnostic} develop efficient algorithms for learning a number of parametric distributions even when a fraction of the data has been arbitrarily corrupted.
These algorithms rely on relatively simple primitives and can thus be efficiently implemented even for high-dimensional distributions~\citepp{diakonikolas2017being}.
In an orthogonal direction, \citett{gao2018robust} draw a connection between robust estimation and GANs~\citepp{goodfellow2014generative} that allows one to approximate complex robust estimators efficiently~\citepp{gao2020generative}.

While these algorithms focus on estimating specific statistical quantities of the data, other work optimizes general notions of risk in the presence of outliers.
For instance, \citett{charikar2017learning} propose algorithms for risk minimization based on recovering a list of possible models and (optionally) using a small uncorrupted dataset to choose between them.
\citett{steinhardt2017certified} focus on the binary classification setting and remove data points that are far from their respective class centroids (measured directly in input space or after projecting data onto the line between the two centroids).
Similarly, \citett{diakonikolas2019sever} and \citett{prasad2018robust} adapt tools from robust mean estimation to robustly estimate the average risk gradient over a (potentially corrupted) dataset.
At a high level, these approaches provide the following theoretical guarantee: if a single data point has a large effect on the model, it will be identified as an outlier.
Such a guarantee prevents an adversary from significantly changing the behavior of the model by just injecting a few inputs.
We refer the reader to \citett{li2018principled}, \citett{steinhardt2018robust}, and \citett{diakonikolas2019recent} for additional references on this line of work.

From the perspective of defending against poisoning on modern ML datasets, both \citett{steinhardt2017certified} and \citett{diakonikolas2019sever} show promising results for regression and classification settings.
Similarly, \citett{paudice2018detection} propose a data pre-filtering approach with outlier detection for linear classifiers. They split a trusted training dataset by class and then train one distance-based outlier detector for each class. When a new untrusted dataset is used for re-training, the outlier detectors remove samples that exceed some score threshold.  In another work, \citett{paudice2018label} mitigate label flipping attacks by using $k$-Nearest-Neighbors ($k$-NN) to re-label each data point in the training set. Specifically, they re-label each data point with the most common label among its $k$ nearest neighbors.
It is worth noting that certain outlier-based defenses can be bypassed by adaptive attacks. Specifically, \citett{koh2018stronger} design attacks that fool anomaly detectors by placing poisoned inputs close to each other and rephrasing poisoning attacks as constrained optimization problems to evade detection.

\subsubsection{Latent Space Signatures}
While input-space outlier detection is simple and intuitive, it is only effective on simple, low-dimensional input domains. In more complex domains, for example image or text data, directly comparing raw input data may not convey any meaningful notion of similarity. Thus, recent work has focused on detecting outliers based on the latent embedding
of a deep neural network. The intuition behind this approach is that latent embeddings capture the signal necessary for classification, thereby making the difference between clean and poisoned inputs more pronounced.

Several ways of analyzing latent model representations arise from this intuition. \citett{tran2018spectral} use tools from robust mean estimation \citepp{diakonikolas2019robust}{lai2016agnostic} to find directions along which the covariance of the feature representations is significantly skewed.  Measuring variation along these directions yields better detection of standard backdoor attacks (Section \ref{sec:backdoor}) than simpler metrics such as the $\ell_2$-distance in feature space. The detection algorithm NIC \citepp{ma2019nic} approximates the distribution of neuron activation patterns and detects inputs that contain the trigger by comparing their activations to this approximate distribution. \citett{chen2018activationclustering} apply clustering algorithms to the latent representations and identify clusters whose members, when removed from training, would be labeled differently by the learned model. \citett{peri2019deepknn} observe that the deep features of poison inputs often lie near the distribution of the target class as opposed to near the distribution of other data with the same label. They use this observation to detect poison examples in clean-label data poisoning attacks. \citett{koh2017influencefunction} use the latent embedding of a model to compute influence functions which measure the effect of each training point on test set performance.
They find that these influence estimates are effective at flagging potentially mislabelled examples (e.g. label flipping attacks) for manual inspection.

\subsubsection{Prediction Signatures}
There is also a number of approaches that directly study the behavior of a model in an end-to-end fashion.
STRIP \citepp{gao2019strip} detects whether a given input contains a backdoor trigger by mixing it with other benign inputs and analyzing the model prediction. The authors posit that if the prediction does not change often, then the model must be heavily relying on a small part of that input. This approach allows the detection of backdoor triggers in deployed models. SentiNet \citepp{chou2020sentinet} uses Grad-Cam \citepp{selvaraju2017grad}, an input saliency mapping method, to pinpoint the features of the input that are most responsible for the model's prediction. If the model only relies on a small part of the input, then it is likely to be relying on a backdoor trigger for its prediction.

\subsection{Identifying Poisoned Models}
The detection approaches described above rely on access to poisoned data used during training.
Thus, they cannot be applied in cases where the entire model training process is outsourced.
There are, however, several defenses that can detect a poisoned model without access to the poisoned training data.

\subsubsection{Trigger Reconstruction}
\label{sec:reconstruction}
One family of approaches aims to recover the backdoor trigger from the model alone \citepp{wang2019neural}{ijcai2019DeepInspect}{guo2019tabor}.
These methods utilize adversarial perturbations to move data towards different target classes. Backdoored models are trained to assign an adversarial label when only a small number of pixels are manipulated to introduce the trigger.  As a result, swapping an image to an adversarial label should require a smaller perturbation than swapping to a non-adversarial label. The backdoor trigger is thus possibly recovered by computing adversarial perturbations for the target labels, and then selecting the smallest perturbation out of all labels.


Neural Cleanse \citepp{wang2019neural}, the first approach to use this observation, is able to detect poisoned models without access to the poisoned dataset. This method does require a number of clean image samples and full access to the trained model parameters with which it can perform gradient descent to find potential triggers.
DeepInspect \citepp{ijcai2019DeepInspect} improves this methodology in three ways. First, it simultaneously recovers potential triggers for multiple classes at once (and hence avoids the computational cost of constructing a potential trigger for each class individually). Second, it relies on model inversion \citepp{fredrikson2015modelinversion} to recover a substitute training dataset, thereby not requiring any clean data. And third, it trains a conditional GAN \citepp{goodfellow2014generative} to estimate the probability density function of potential triggers for any target class. Another trigger reconstruction defense called TABOR \citepp{guo2019tabor} further improves upon Neural Cleanse by enhancing the fidelity of the reconstructed backdoor triggers via heuristic regularization methods.
Most recently, \citett{wang2020practical} study the data-limited (one shot per class) and data-free cases.
In the data-limited case, they reconstruct a universal adversarial perturbation (i.e. trigger) and the image-wise perturbations for each label, the similarity of which is then used for backdoor detection. If a model is backdoored, then the universal perturbation and the image-wise one for the backdoor target label may share strong similarities due to the existence of a backdoor shortcut. 
In the data-free case, they first generate perturbed images from random seed images by maximizing neuron activations. Then, they detect if a model is backdoored by investigating the magnitude of the change in logit outputs with respect to random images and perturbed images.

\subsubsection{Trigger-agnostic Detection}
Different from the detection pipelines discussed above, MNTD \citepp{xu2019Meta}  predicts whether a model is backdoored by examining its behavior on carefully crafted inputs. MNTD first generates a battery of benign and backdoored models. Then, a query set of images is generated and pushed through the battery of networks to create outputs.  Finally, a binary meta-classifier is trained to examine these outputs and determine whether a model is backdoored. 
The query set is then jointly optimized with the parameters of the meta-classifier to obtain a high accuracy meta-classifier. Interestingly, this approach appears to detect  attacks on architectures outside of the ensemble used to train the meta-classifier \citepp{xu2019Meta}.
\citett{huang2020one} define a ``one-pixel'' signature of a network, which is the collection of single-pixel adversarial perturbations that most effectively impact the label of a collection of images.  They train a meta-classifier on these signatures to classify backdoored vs clean models.  
\citett{kolouri2020universal} present a method for examining how networks respond to trigger-like patterns. A set of ``universal litmus patterns'' are pushed through a network, and a meta-classifier is trained on the resulting logits to determine whether the network is backdoored \citepp{kolouri2020universal}. Importantly, this method seems to generalizes well across architectures and backdoor attacks.

\subsection{Repairing Poisoned Models after Training}
While detection is useful for defending against data poisoning attacks, the methods described above only indicate whether or not an attack has occurred. Another class of methods removes backdoors from an already trained model without re-training the model from scratch.
We describe methods that rely on (approximate) knowledge of the trigger and also methods that do not require such knowledge.

\subsubsection{Patching Known Triggers}
One line of defense strategies relies on trigger reconstruction methods to recover an approximation of the injected trigger as seen in Section \ref{sec:reconstruction}.
Once the trigger is identified, there are multiple ways of rendering it inactive.
Neural Cleanse \citepp{wang2019neural} shows that it is possible to identify which neurons are strongly activated by the presence of the trigger and use them to detect inputs containing the trigger during testing.
Similarly, they remove the influence of these neurons by pruning them.
Additionally, Neural Cleanse fine-tunes the model to \emph{unlearn} the trigger by adding it to clean samples and training with correct labels. 
Rather than reconstructing a single trigger, the authors of \citett{ijcai2019DeepInspect}, \citett{qiao2019defending}, and \citett{zhu2020gangsweep} model a distribution of possible triggers using a GAN, which is then sampled to train an immunized model.

\subsubsection{Trigger-agnostic Backdoor Removal}
Another way to remove backdoor behavior is to modify the model, only keeping the parts necessary for the intended tasks.  Since the backdoor is not active during forward passes on clean data, the backdoor behavior can be removed during this process.
\citett{liu2018fine} attempt such a defense by pruning neurons that are dormant---i.e., they are not activated on clean inputs.
Specifically, their defense tests the poisoned model with clean inputs, records the average activation of each neuron, and iteratively prunes neurons in increasing order of average activation. 
However, this pruning defense cannot remove the backdoor without significantly degrading performance \citepp{liu2018fine}.

To overcome the performance loss of pruning defenses alone, several works propose fine-tuning the model on a clean dataset \citepp{liu2018fine}{chen2019refit}{liu2020removing}.
Since the dataset does not include the backdoor trigger, the backdoor behavior may eventually be forgotten after updating the parameters during fine-tuning. 
Combining pruning with fine-tuning can indeed remove backdoors while preserving the overall model accuracy, even when an adversary crafts pruning-aware attacks \citepp{liu2018fine}.
However, if the dataset used for fine-tuning is small, model performance may suffer significantly \citepp{chen2019refit}.


To better preserve the model's accuracy on clean data, the watermark removal framework REfiT \citepp{chen2019refit} leverages elastic weight consolidation \citepp{kirkpatrick2017overcoming}. This defensive training process slows down the learning of model weights that are relevant to the main prediction task while updating other weights that are likely responsible for memorizing watermarks. Similarly, WILD \citepp{liu2020removing} includes a feature distribution alignment scheme to achieve a similar goal.

\subsection{Preventing Poisoning during Training}
The methods described above aim to either detect or fix an already poisoned model.
Here, we describe training-time strategies to avoid backdoor injection in the first place.

\subsubsection{Randomized Smoothing}
Randomized smoothing \citepp{lecuyer2019certifieddp}{cohen2019certified} was originally proposed to defend against evasion attacks \citepp{biggio2013evasion}.
Starting with a base model, a smoothed version of that model is defined by replacing the model prediction on each data point by the majority prediction in its neighborhood.
The outputs of the smoothed model can be computed efficiently, while at the same time certifying its robustness to input perturbations \citepp{cohen2019certified}.

In the context of data poisoning, the goal is to protect the model against perturbations to the training set.
Thus, the goal of \emph{certifiable robustness} is for each test point, to return a prediction as well as a certificate that the prediction would not change had some quantity of training data been modified.
To do so, one can view the entire training-plus-single-prediction pipeline as a function which can be robustified against input perturbations \citepp{weber2020rab}{ rosenfeld2020certified}.
\citett{weber2020rab} apply this defense to backdoor attacks, where backdoor pixel perturbations are in the continuous space. On the other hand, \citett{rosenfeld2020certified} use this defense against label flipping attacks where label perturbations are in the discrete space.

\subsubsection{Majority Vote Mechanisms}
A number of approaches utilize majority vote mechanisms to ignore poisoned samples.
The underlying assumption is that the number of poisoned samples injected by the attacker is small compared to the size of the overall training dataset. Therefore, the poison samples will not significantly influence the majority vote when voters each use only a subset of the data.
For example, Deep Partition Aggregation \citepp{levine2020deeppartition} learns multiple base classifiers by partitioning the training set into disjoint subsets.
Similarly, \citett{jia2020intrinsic} train multiple base models on random subsamples of the training dataset.
The base models are then combined via majority vote to produce an aggregate model, the robustness of which can be confirmed empirically and certified theoretically.\citett{jia2020certified} predict the label of a test example via majority vote among the labels of its $k$ nearest neighbors or all of its neighbors within radius $r$ in the training dataset.

\subsubsection{Differential Privacy} 
Differential Privacy (DP) \citepp{dwork2006calibrating} was originally designed to protect the privacy of individuals contributing data. The core idea is that if the output of the algorithm remains essentially unchanged when one individual input point is added or subtracted, the privacy of each individual is preserved.
From the perspective of data poisoning, differential privacy ensures that model predictions do not depend too much on individual data points.
Thus, models will not be disproportionately affected by poisoned samples.
\citett{ma2019dplearners} study defenses based on DP against data poisoning from the practical and theoretical perspectives.
\citett{hong2020GradientShaping} empirically show that the off-the-shelf mechanism DP-SGD \citepp{abadi2016dpsgd}, which clips and noises gradients during training, can serve as a defense. They point out that the main artifacts of gradients computed in the presence of poisoning are that their $\ell_2$-norms have higher magnitudes and their orientation differs from clean gradients. Since DP-SGD bounds gradient magnitudes by clipping and minimizes the difference in orientation by random noise addition, it is successful in defending against poisoning attacks \citepp{hong2020GradientShaping}.  A recent line of work proves that data augmentations which yield strong empirical defense against poisoning and backdoor attacks also enhances differential privacy guarantees all without degrading model performance \citepp{borgnia2020strong} { borgnia2021dp}.

\subsubsection{Input Preprocessing}
Additionally, some works propose modifying the model input, during training or testing, to prevent the model from recognizing the backdoor trigger \citepp{liu2017neural}{ borgnia2020strong}.
\citett{liu2017neural} utilize an autoencoder \citepp{vincent2008extracting} trained on clean data to preprocess the input. Since the input is not perfectly reconstructed (especially given that the autoencoder is not particularly sensitive to the trigger), the model is unlikely to recognize the trigger.  \citett{borgnia2020strong} and \citett{borgnia2021dp} propose using strong data augmentations during training, such as \emph{mixup} \citepp{zhang2017mixup}, CutMix \citepp{yun2019cutmix}, and MaxUp \citepp{gong2020maxup}. Dramatic data augmentations sufficiently disrupt triggers and perturbations to the training data, foiling the attack. This approach is highly effective against both training-only and backdoor attacks and has the added benefit that it does not degrade model performance \citepp{borgnia2020strong} { borgnia2021dp}.  Adversarial training, which is typically used to defend against evasion attacks, has also been adapted to defend against training-only and backdoor train-time attacks \citepp{geiping2021doesn}.  This work trains against intentionally poisoned data in order to desensitize a model to dataset manipulations.

\subsection{Defenses for Federated Learning}
In Federated Learning (FL), a global model is trained by utilizing local data from many clients, providing a venue for new poisoning attacks from malicious clients.
Since this setting allows for fundamentally different attacks compared to other learning settings (cf. Section~\ref{sec:poisoning-fl}), a number of application-specific defenses have been developed for FL.
These include robust federated aggregation algorithms, robust federated training protocols, and post-training measures. 

\subsubsection{Robust Federated Aggregation}
Robust federated aggregation algorithms attempt to nullify the effects of attacks while aggregating client updates.  These methods can be broadly classified into two types; one type identifies and down-weights the malicious updates, while a second
type does not attempt to identify malicious clients and instead computes aggregates in a way which is resistant to poisons.  A prototypical idea for this second method estimates a true ``center’’ of the received model updates rather than taking a weighted average. 

\citett{fung2018mitigating} identify sybils-based attacks, including label flipping and backdoors, as group actions. Sybils share a common adversarial objective whose updates are more similar to each other than honest clients. They propose FoolsGold, which calculates the cosine similarity of the gradient updates from clients, reduces aggregation weights of clients that contribute similar gradient updates, thus promoting contribution diversity.
Another avenue for defense in the FL setting is learning-based robust aggregation.
Specificically,
\citett{li2020learning} utilize a variational autoencoder (VAE)~\citepp{kingma2013auto} to detect and remove malicious model updates in aggregation, where the VAE is trained on clean model updates. The encoder projects model updates into low-dimensional embeddings in latent space, and the decoder reconstructs the sanitized model updates and generates a reconstruction error. 
The idea is that the low-dimensional embeddings retain essential features, so malicious updates are sanitized and trigger much higher reconstruction errors while benign updates are unaffected. Updates with higher reconstruction errors are deemed malicious and are excluded from aggregation.
This method assumes that a clean dataset is available to train the detection model. 

Another approach involves Byzantine-robust aggregation techniques which resist manipulation, even without identifying malicious clients. 
Two such algorithms, called Krum and Multi-Krum \citepp{blanchard2017machine}, select representative gradient updates which are close to their nearest neighbor gradients. Another algorithm called Bulyan \citepp{mhamdi2018hidden} first uses another aggregation rule such as Krum to iteratively select benign candidates and then aggregates these candidates by a variant of the trimmed mean \citepp{yin2018byzantine}.

Another class of approaches employs the coordinate-wise median gradient \citepp{yin2018byzantine}, geometric median of means \citepp{chen2017distributed}, and approximate geometric median \citepp{pillutla2019robust} since median-based computations are more resistant to outliers than mean-based aggregation. These methods are effective in robust distributed learning where the data of each user comes from the same distribution. However, they are less effective in FL where local data may be collected in a non-identically distributed manner across clients. Also, these methods are shown, both theoretically and empirically, to be less effective in the high-dimensional regime \citepp{mhamdi2018hidden}. Another Byzantine-robust method called RSA \citepp{li2019rsa} penalizes parameter updates which move far away from the previous parameter vector and provides theoretical robustness guarantees. Unlike earlier methods, RSA does not assume that all workers see i.i.d. data from the same distribution \citepp{li2019rsa}. Alternatively, \citett{fu2019attack} estimate a regression line by a repeated median estimator \citepp{siegel1982robust} for each parameter dimension of the model updates, then dynamically assign aggregation weights to clients based on the residual of their model updates to that regression line.

\subsubsection{Robust Federated Training}
In addition to robust federated aggregation, several FL protocols mitigate  poisoning attacks during training.  \citett{sun2019can} show that clipping the norm of model updates and adding Gaussian noise can mitigate backdoor attacks that are based on the model replacement paradigm \citepp{bagdasaryan2020backdoor}{bhagoji2019analyzing}. This work highlights the success of this method on the realistic federated EMNIST dataset \citepp{cohen2017emnist}{caldas2018leaf}.

\citett{andreina2020baffle} leverage two specific features of federated learning to defend against backdoor attacks. Specifically, they utilize global models produced in the previous rounds and the fact that the attacker does not have access to a substantial amount of training data. They propose BaFFLe \citepp{andreina2020baffle}, which incorporates an additional validation phase to each round of FL.
That is, a set of randomly chosen clients validate the current global model by computing a validation function on their private data and report whether the current global model is poisoned. The server decides to accept or reject the global model based on the feedback from validating clients. Specifically, the validation function compares the class-specific misclassification rates of the current global model with those of the accepted global models in the previous rounds and raises a warning when the misclassification rates differ significantly, which may indicate backdoor poisoning. 

\subsubsection{Post-Training Defenses}
Other defense strategies focus on restoring the poisoned global model after training. \citett{wu2020mitigating} extend pruning and fine-tuning methods \citepp{liu2018fine} to the FL setting to repair backdoored global models. Their method requires clients to rank the dormant level of neurons in the neural network using their local data, since the server itself cannot access clean training data. They then select and remove the most redundant neurons via majority vote \citepp{wu2020mitigating}.

\subsection{Open Problems}
\begin{itemize}
    \item \textbf{Defenses beyond image classification}:  Although data poisoning and backdoor attacks have been applied in a variety of domains, image classification remains to be the major focus of research for defenses. It is thus crucial that these defenses are applied to other domains in order to understanding potential for real-world use, as well as any shortcomings.
    
    \item \textbf{Navigating trade-offs between accuracy, security, and data privacy:} 
    Modern large-scale ML systems strive to achieve high accuracy while maintaining user privacy.
    However, these goals seem to be at odds in the presense of data poisoning.
    In fact, most FL defenses rely on direct access to model updates, which can reveal information about user data \citepp{zhu2019deep}{geiping2020inverting}. Achieving security against poisoning while maintaining accuracy and privacy appears to be elusive given our current methods.
    
    \item \textbf{Can defenses be bypassed without access to training?} \citett{tan2019bypassing} show that one can bypass certain outlier-based defenses 
    by enforcing that the internal representations of a model on poison examples during training are similar to those corresponding to clean examples. The open question is whether these defenses can be bypassed \emph{without access to the training protocol}.
    \item \textbf{Efficient and practical defenses}: Many approaches to identifying poisoned models require producing a set of auxiliary clean and poisoned models to train a detector \citepp{xu2019Meta}{huang2020one}{kolouri2020universal}, but this process is computational expensive. Moreover, generating additional models for trigger-agnostic methods or reconstructing possible triggers in trigger reconstruction methods \citepp{wang2019neural}{guo2019tabor} requires a clean dataset, which may not be feasible in practice. Therefore, designing efficient and practical defense methods with less data and computation requirements is essential for practical application.
    \item \textbf{Differential privacy and data poisoning:} \citett{hong2020GradientShaping} and \citett{jagielski2020auditing} show that there remains a massive gap between the theoretical worst-case lower bounds provided by DP mechanisms and the empirical performance of the defenses against data poisoning. It is however unclear if this gap is due to existing attacks being insufficient or due to the theoretical bounds being unnecessarily pessimistic.
    \item \textbf{Certified defenses against poisoning attacks:} 
    Certified defenses against poisoning attacks are still far from producing meaningful guarantees in realistic, large-scale settings.
    Moreover, they are particularly hard to study in a federated learning setting.
    Instead of analyzing the influence of training datasets on model predictions in an end-to-end manner like in the centralized setting \citepp{weber2020rab}{rosenfeld2020certified}, one needs to consider how the local training datasets influence the local updates, and how the performance of the global model is influenced by these local updates through aggregation.
    
    \item \textbf{Detection of inconspicuous poison examples:} Defense strategies rely on poison examples or backdoor model behavior being noticeably irregular in the context of the ambient dataset. Detecting malicious behavior when it does not appear anomalous is a much harder task, and existing methods often fail.  Similarly, anomaly detection is ineffective in federated learning, where each client may have a dramatically different underlying data distribution.  Picking out malicious clients from benign, yet atypical ones is an important open problem.
\end{itemize}

\section{Conclusion}
\label{discussion}
The expansion of the power and scale of machine learning models has been accompanied by a corresponding expansion of their vulnerability to attacks.  In light of the rapid surge in work on data poisoning and defenses against it, we provide a birds-eye perspective by systematically dissecting the wide range of research directions within this space.  The open problems we enumerate reflect the interests and experience of the authors and are not an exhaustive list of outstanding research problems.  Moving forward, we anticipate work not only on these open problems but also on datasets and benchmarks for comparing existing methods since controlled comparisons are currently lacking in the data poisoning and backdoor literature.  We hope that the overhead perspective we offer helps to illuminate both urgent security needs within industry and also a need to understand security vulnerabilities so that the community can move towards closing them.

\section*{Acknowledgements}
We thank Jiantao Jiao, Mohammad Mahmoody, and Jacob Steinhardt for helpful pointers to relevant literature.

Multiple contributors to this work were supported by the Defense Advanced Research Projects Agency (DARPA) GARD, QED4RML and D3M programs.  Additionally, support for Li and Xie was provided by
NSF grant CCF-1910100 and the Amazon research award program.  Song and Chen were supported by 
NSF grant TWC-1409915, Berkeley DeepDrive, and the Facebook PhD Fellowship. 
Madry and Tsipras were supported by
NSF grants CCF-1553428, CNS-1815221, and the Facebook PhD Fellowship.
Goldstein, Goldblum, and Schwarzschild were supported by NSF grant DMS-1912866, the ONR MURI Program, and the Sloan Foundation.

\bibliography{refs}

\end{document}